\documentclass[11pt,a4paper]{article}
\usepackage[hyperref]{ranlp2023}
\usepackage{times}
\usepackage{latexsym}
\usepackage{geometry}%
\usepackage{multicol}
\usepackage{graphicx}%
\usepackage{booktabs}

\usepackage{amsmath}

\usepackage{microtype}

\aclfinalcopy 

\title{Cordyceps@LT-EDI : Depression Detection with Reddit and Self-training}

\author{Dean Ninalga \\
\texttt{justin.ninalga@mail.utoronto.ca}}

\date{}

\begin{document}
\maketitle
\begin{abstract}

 Depression is debilitating, and not uncommon. 
 Indeed, studies of excessive social media users show correlations with depression, ADHD, and other mental health concerns. 
Given that there is a large number of people with excessive social media usage, then there is a significant population of potentially undiagnosed users and posts that they create.
 In this paper, we propose a depression severity detection system using a semi-supervised learning technique to predict if a post is from a user who is experiencing severe, moderate, or low (non-diagnostic) levels of depression. Namely, we use a trained model to classify a large number of unlabelled social media posts from Reddit\footnote{https://www.reddit.com/}, then use these generated labels to train a more powerful classifier.
We demonstrate our framework on \emph{Detecting Signs of Depression from Social Media Text - LT-EDI@RANLP 2023} \cite{Sampath-depression-2023-overview} shared task, where our framework ranks  3rd overall.
\end{abstract}

\section{Introduction}

\subsection{Depression and Social Media}
A unique feature of depression is its effect on cognitive and verbal patterns. 
For example, depression diagnosis is correlated to the frequency of personal pronoun usage and the usage of positive-negative words \cite{Edwards2017AMO, Tlbll2019LinguisticFI}.
Additionally, persons suffering from depression often connect vicious yet potentially fictional narratives to benign experiences, generally increasing the number of overwhelming situations they may experience \cite{Kanter2008TheNO}. 
People may then go to social media and online forums like Reddit to discuss and post about traumatizing experiences and may publicly reflect on their thoughts and behavior. It is unsurprising, therefore, to find a wealth of attempts (as surveyed by \citet{Hasib2023DepressionDF}) to use social media posts to create a potential diagnostic screening tool through language modeling. 
Recently, language models can accurately predict symptoms before practitioners record them \cite{Eichstaedt2018FacebookLP, Reece2016ForecastingTO}.

There are significant challenges to data collection in the depression detection setting despite a potential abundance of data that likely exists on social media.
Indeed, excessive social media usage itself correlates with depression, ADHD, and other serious mental health diagnoses \cite{Hussain2019TheAB}. However, \citet{Guntuku2017DetectingDA} observe that most attempts at data collection rely on a self-declaration or a past diagnosis of depression, allowing for the possibility of non-actively depressed individuals creating depression-positive data.
In this paper, we will attempt to apply an automatic data collection process from social media through a semi-supervised approach. 

\subsection{Background on Self-training}
Self-training techniques \cite{Scudder1965ProbabilityOE} are 
 a type of semi-supervised learning and are well known in various areas of research (e.g. \citet{Zoph2020RethinkingPA, Xie2019SelfTrainingWN, Sahito2021BetterSF}). 
These techniques in broad terms, take a trained model, generate labels for a large set of unlabeled data, then train a new model incorporating the clean labels, generated labels, and unlabeled data.
Where the new model is typically of the same size, or bigger, as the original trained model.
Surprisingly, however, little work has been done exploring how to apply this process in the specific case of depression detection across many social media.

To summarize, our main contributions are the following:
\begin{itemize}
    \item We describe our framework based on self-training.
    \item We demonstrate our framework on \emph{Detecting Signs of Depression from Social Media Text - LT-EDI@RANLP 2023} \cite{Sampath-depression-2023-overview} shared task, comparing to recent work.
    \item We describe areas where pseudo-labeling can advance depression detection modeling.
\end{itemize}

\section{Related Work}
Recent work has demonstrated semi-supervised learning techniques using unlabeled Twitter data for depression detection as surveyed by \cite{Zhang2022NaturalLP}.
However, these studies tend to solely rely on Twitter\footnote{https://twitter.com/} data as their source of unlabeled texts \cite{Zhang2022NaturalLP, Yazdavar2017SemiSupervisedAT}.
Here, we will use Reddit for our semi-supervised approach. 

\citet{Poswiata2022OPILTEDIACL2022DS} also uses the \emph{Reddit Mental Health Dataset} \cite{low2020natural} in their depression detection system for last year's iteration of the shared task.
However, \citet{Poswiata2022OPILTEDIACL2022DS} do not generate pseudo-labels but instead use the data for a pre-training task that is specifically designed for depression detection. 
\citet{Pirina2018IdentifyingDO} suggested that the selection of Reddit forums (or \emph{subreddits}) in the training data may influence the quality of classifiers.
Here, our goal is to automate this selection process without having to rely on \emph{subreddit} specific information and rely solely on the posts themselves.

\section{Methodology}
Here we will provide the major implementation details of our solution in this section.
See Table \ref{tab:hyp} for further information on training hyper-parameter details used throughout. 
\subsection{Data Cleaning}
We perform a few basic data-cleaning steps for any samples fed to the classifier. That is, we remove any newline and tab characters, strip leading and trailing white spaces, and replace all links with an identical string.
Additionally, we remove duplicated texts and drop samples in the shared-task training set if it is also contained in the shared-task development set.
In total, we dropped 128 duplicated samples.

\begin{table*}
\centering
\begin{tabular}{lll}
\toprule
\textbf{Name} & Dev & \textbf{Test}\\
\hline
MentalRoBERTa \cite{ji2022mentalbert} + \emph{pl+ft} (ours) & \textbf{0.7407} & 0.4309 \\
MentalRoBERTa \cite{ji2022mentalbert} + \emph{pl} & 0.5359 & 0.3975 \\

\hline
MentalRoBERTa \cite{ji2022mentalbert} & 0.578 & 0.44 \\
MentalXLNet \cite{ji-domain-specific} & 0.5714 & \textbf{0.4443} \\
MentalBERT \cite{ji2022mentalbert} & 0.5648 &  0.3901 \\
\hline
RoBERTa \cite{DBLP:journals/corr/abs-1907-11692} & 0.5627 & 0.3953 \\
BERT \cite{DBLP:journals/corr/abs-1810-04805} & 0.5512 & 0.3981 \\
\bottomrule

\end{tabular}
\caption{\label{tab:res}
\emph{Macro}-averaged F1-Score results on the development and test set of the shared task. The best score on each set is highlighted. The top two rows highlight a single run of our approach: training on only pseudo-labels (\emph{pl}) and then finetuning (\emph{ft}). The next three rows detail the finetuning results of recently released pre-trained models for mental health. In the last two rows, we present a baseline using well-known models.}
\end{table*}

\begin{table}
\centering
\begin{tabular}{l|l}
\toprule
\textbf{Hyper-Parameter} & \textbf{Value}\\
\hline
Optimizer & Adam \\
Learning Rate & 1e-5 \\
Max Input Length & 256 \\
Batch Size & 8 \\
\bottomrule

\end{tabular}
\caption{\label{tab:hyp}
Training Hyper-parameter Details}
\end{table}

\subsection{Pre-Trained Models}
Leveraging pre-trained language representations is a proven way to boost performance on essentially any given NLP task.
Downstream task performance gains are even more prominent if the pre-training task is identical to the downstream tasks and uses large amounts of data.
    To that end, we use MentalRoBERTa \cite{ji2022mentalbert} as our model of choice for training and inference.
MentalRoBERTa \cite{ji2022mentalbert} is a RoBERTa \cite{Liu2019RoBERTaAR} model that is further pre-trained on Reddit mental-health-related data.

\subsection{Self-Training}
The details of our self-training and pseudo-labeling procedure are as follows.
Firstly, we train a teacher model using the annotated training data.
Next, we use the trained teacher model to generate predictions on the unlabeled data: \emph{Reddit Mental Health Dataset} \cite{low2020natural}. 
Here, we want to keep the highest-ranked 30,000 samples with the highest-valued predicted logit for any of the three label categories.
For example, we only include a post in the severe depression category if the teacher model is very confident that a sample belongs in the depression category relative to all other posts.
Subsequently, the resulting 90,000 posts are then assigned pseudo-labels based on the previously assigned groupings, where we assume that each sample belongs to its respective category. Here, we do not consider the categorical probability distribution (as predicted by the teacher) since we are only keeping samples with high confidence. In practice, the predicted output probabilities of the 90,000 posts are very close to 1 for their respective category, hence, using the predicted probabilities adds very little information.
Next, we use the 90,000 posts alongside the pseudo-labels to construct a new dataset which is used to train a new student model. Note, here we use the same model architecture for both the teacher and student.
Finally, the student model is finetuned with the clean training data and then used for inference on the test set.

\section{Experiments}
\subsection{Experimental Setup}
We compare our setup to several other state-of-the-art pre-trained models we finetuned for the shared task. 
We report the macro-averaged F1-Score on the test and development sets. Where report the average score over five runs, unless otherwise stated.
We perform all experiments on a single T4 GPU.

\begin{figure*}[h]
  \includegraphics[width=\textwidth]{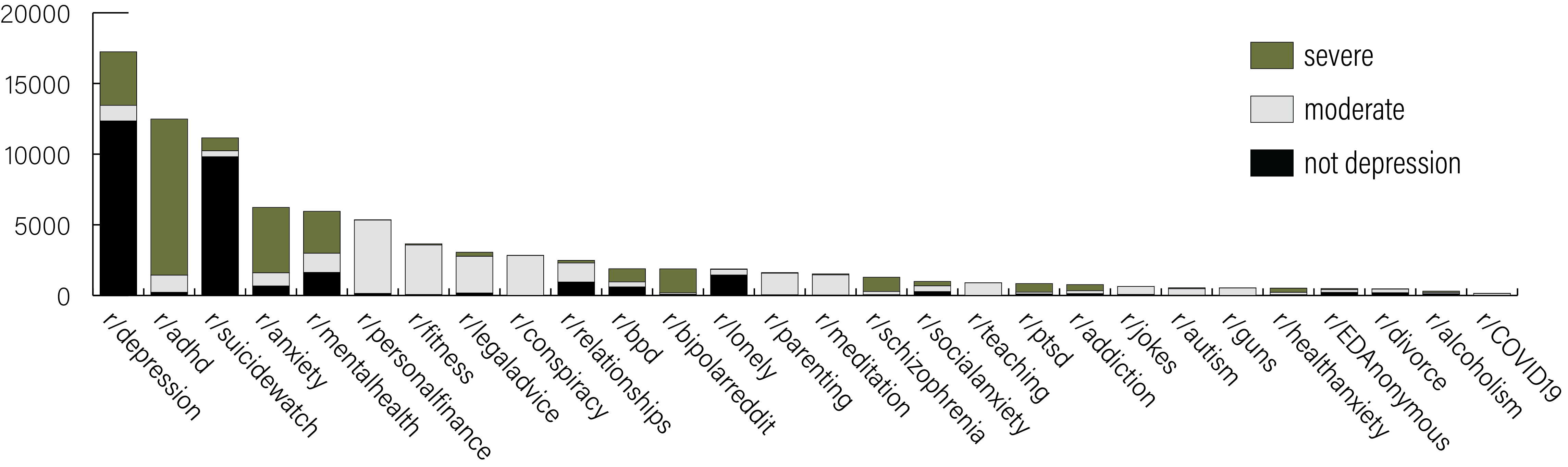}
  \centering
  \caption{\label{fig:sr}
  Breakdown of the pseudo-labels on each subreddit in the \emph{Reddit Mental Health Dataset} \cite{low2020natural}}
\end{figure*}

\subsection{Results}
We present our full results in Table \ref{tab:res}.
Indeed, our complete approach of self-training with MentalRoBERTa \cite{ji2022mentalbert} performs the best on the development set by a wide margin. However, our approach performs narrowly worse than MentalXLNet \cite{ji-domain-specific} on the test set.
Given this disparity in development and test set performance, future work should explore regulation techniques (e.g. augmentation and ensembling methods) to accompany the self-training approach. Nonetheless, our approach still places 3rd overall in the shared task. 

\section{Exploratory analysis}
We present an analysis of our generated pseudo-labels on the \emph{Reddit Mental Health Dataset} \cite{low2020natural}.
Recall, that we assign a pseudo-label to a post only if the post is ranked in the top 30,000 in any of the three depression severity labels.   
In Figure \ref{fig:sr} we break down the distribution of the labels across the sources of these labels. 
Notably, we find about 60\% of our generated labels are contained in five subreddits: \emph{`r/depression'}, \emph{`r/adhd'} \emph{`r/suicidewatch'}, \emph{`r/anxiety'}, \emph{`r/mentalhealth'}.  In particular, the subreddit \emph{`r/adhd'} hosts the most pseudo-labels in the \emph{severe} category out of any subreddit by a wide margin,  accounting for about 37\% of all pseudo-labels in the category.

There are multiple explanations for the above findings. 
Indeed, ADHD can co-occur with depression and can be seen as an early indication of a future depression diagnosis \cite{Meinzer2017ADHDAT}.
Additionally, ADHD and depression have overlapping symptoms \cite{Riglin2020ADHDAD}. 
Thus, it is possible that there is some level of overlapping language or similar verbal processes shared between the two disorders.
We encourage future work to explore alternative explanations and leverage this connection between ADHD and depression in the depression-detection setting.

\section{Conclusion}
In this paper, we present our framework based on self-training and demonstrate its performance on the \emph{Detecting Signs of Depression from Social Media Text - LT-EDI@RANLP 2023} \cite{Sampath-depression-2023-overview} shared task. 
Given the disparities observed in the development set and test set F1-score performance, future work should explore regulation techniques (e.g. augmentation and ensembling methods) to accompany the self-training approach. Nonetheless, our approach still places 3rd overall in the shared task. 

With our use of pseudo-labeling on Reddit, we highlighted ADHD-focused forums as a major source of (non-diagnostic) severe depression classifications and discussed some explanations. 
We hope our work serves as a starting point for further investigation of the linguistic patterns of depression overlapping with other mental disorders.

\bibliographystyle{acl_natbib}
\bibliography{ranlp2023.bib}

\begin{thebibliography}{25}
\expandafter\ifx\csname natexlab\endcsname\relax\def\natexlab#1{#1}\fi

\bibitem[{Devlin et~al.(2018)Devlin, Chang, Lee, and
  Toutanova}]{DBLP:journals/corr/abs-1810-04805}
Jacob Devlin, Ming{-}Wei Chang, Kenton Lee, and Kristina Toutanova. 2018.
\newblock \href {http://arxiv.org/abs/1810.04805} {{BERT:} pre-training of deep
  bidirectional transformers for language understanding}.
\newblock \emph{CoRR}, abs/1810.04805.

\bibitem[{Edwards and Holtzman(2017)}]{Edwards2017AMO}
T.~Edwards and Nicholas~S. Holtzman. 2017.
\newblock A meta-analysis of correlations between depression and first person
  singular pronoun use.
\newblock \emph{Journal of Research in Personality}, 68:63--68.

\bibitem[{Eichstaedt et~al.(2018)Eichstaedt, Smith, Merchant, Ungar, Crutchley,
  Preotiuc-Pietro, Asch, and Schwartz}]{Eichstaedt2018FacebookLP}
Johannes~C. Eichstaedt, Robert~J. Smith, Raina~M. Merchant, Lyle~H. Ungar,
  Patrick Crutchley, Daniel Preotiuc-Pietro, David~A. Asch, and H.~A. Schwartz.
  2018.
\newblock Facebook language predicts depression in medical records.
\newblock \emph{Proceedings of the National Academy of Sciences of the United
  States of America}, 115:11203 -- 11208.

\bibitem[{Guntuku et~al.(2017)Guntuku, Yaden, Kern, Ungar, and
  Eichstaedt}]{Guntuku2017DetectingDA}
Sharath~Chandra Guntuku, David~Bryce Yaden, Margaret~L. Kern, Lyle~H. Ungar,
  and Johannes~C. Eichstaedt. 2017.
\newblock Detecting depression and mental illness on social media: an
  integrative review.
\newblock \emph{Current Opinion in Behavioral Sciences}, 18:43--49.

\bibitem[{Hasib et~al.(2023)Hasib, Islam, Sakib, Akbar, Razzak, and
  Alam}]{Hasib2023DepressionDF}
Khan~Md Hasib, Md~Rafiqul Islam, Shadman Sakib, Md.~Ali Akbar, Imran Razzak,
  and Mohammad~Shafiul Alam. 2023.
\newblock Depression detection from social networks data based on machine
  learning and deep learning techniques: An interrogative survey.
\newblock \emph{IEEE Transactions on Computational Social Systems}.

\bibitem[{Hussain and Griffiths(2019)}]{Hussain2019TheAB}
Zaheer Hussain and Mark~D. Griffiths. 2019.
\newblock The associations between problematic social networking site use and
  sleep quality, attention-deficit hyperactivity disorder, depression, anxiety
  and stress.
\newblock \emph{International Journal of Mental Health and Addiction}, 19:686
  -- 700.

\bibitem[{Ji et~al.(2022)Ji, Zhang, Ansari, Fu, Tiwari, and
  Cambria}]{ji2022mentalbert}
Shaoxiong Ji, Tianlin Zhang, Luna Ansari, Jie Fu, Prayag Tiwari, and Erik
  Cambria. 2022.
\newblock {MentalBERT: Publicly Available Pretrained Language Models for Mental
  Healthcare}.
\newblock In \emph{Proceedings of LREC}.

\bibitem[{Ji et~al.(2023)Ji, Zhang, Yang, Ananiadou, Cambria, and
  Tiedemann}]{ji-domain-specific}
Shaoxiong Ji, Tianlin Zhang, Kailai Yang, Sophia Ananiadou, Erik Cambria, and
  J{\"o}rg Tiedemann. 2023.
\newblock \href {https://arxiv.org/abs/2304.10447} {Domain-specific continued
  pretraining of language models for capturing long context in mental health}.
\newblock \emph{arXiv preprint arXiv:2304.10447}.

\bibitem[{Kanter et~al.(2008)Kanter, Busch, Weeks, and
  Landes}]{Kanter2008TheNO}
Jonathan~W. Kanter, Andrew~M. Busch, Cristal~E. Weeks, and Sara~J. Landes.
  2008.
\newblock The nature of clinical depression: Symptoms, syndromes, and behavior
  analysis.
\newblock \emph{The Behavior Analyst}, 31:1--21.

\bibitem[{Liu et~al.(2019{\natexlab{a}})Liu, Ott, Goyal, Du, Joshi, Chen, Levy,
  Lewis, Zettlemoyer, and Stoyanov}]{DBLP:journals/corr/abs-1907-11692}
Yinhan Liu, Myle Ott, Naman Goyal, Jingfei Du, Mandar Joshi, Danqi Chen, Omer
  Levy, Mike Lewis, Luke Zettlemoyer, and Veselin Stoyanov. 2019{\natexlab{a}}.
\newblock \href {http://arxiv.org/abs/1907.11692} {Roberta: {A} robustly
  optimized {BERT} pretraining approach}.
\newblock \emph{CoRR}, abs/1907.11692.

\bibitem[{Liu et~al.(2019{\natexlab{b}})Liu, Ott, Goyal, Du, Joshi, Chen, Levy,
  Lewis, Zettlemoyer, and Stoyanov}]{Liu2019RoBERTaAR}
Yinhan Liu, Myle Ott, Naman Goyal, Jingfei Du, Mandar Joshi, Danqi Chen, Omer
  Levy, Mike Lewis, Luke Zettlemoyer, and Veselin Stoyanov. 2019{\natexlab{b}}.
\newblock Roberta: A robustly optimized bert pretraining approach.
\newblock \emph{ArXiv}, abs/1907.11692.

\bibitem[{Low et~al.(2020)Low, Rumker, Torous, Cecchi, Ghosh, and
  Talkar}]{low2020natural}
Daniel~M Low, Laurie Rumker, John Torous, Guillermo Cecchi, Satrajit~S Ghosh,
  and Tanya Talkar. 2020.
\newblock Natural language processing reveals vulnerable mental health support
  groups and heightened health anxiety on reddit during covid-19: Observational
  study.
\newblock \emph{Journal of medical Internet research}, 22(10):e22635.

\bibitem[{Meinzer and Chronis-Tuscano(2017)}]{Meinzer2017ADHDAT}
Michael~C. Meinzer and Andrea Chronis-Tuscano. 2017.
\newblock Adhd and the development of depression: Commentary on the prevalence,
  proposed mechanisms, and promising interventions.
\newblock \emph{Current Developmental Disorders Reports}, 4:1--4.

\bibitem[{Pirina and Çagri Ç{\"o}ltekin(2018)}]{Pirina2018IdentifyingDO}
Inna~Loginovna Pirina and Çagri Ç{\"o}ltekin. 2018.
\newblock Identifying depression on reddit: The effect of training data.
\newblock In \emph{Conference on Empirical Methods in Natural Language
  Processing}.

\bibitem[{Poswiata and Perelkiewicz(2022)}]{Poswiata2022OPILTEDIACL2022DS}
Rafal Poswiata and Michal Perelkiewicz. 2022.
\newblock Opi@lt-edi-acl2022: Detecting signs of depression from social media
  text using roberta pre-trained language models.
\newblock In \emph{LTEDI}.

\bibitem[{Reece et~al.(2016)Reece, Reagan, Lix, Dodds, Danforth, and
  Langer}]{Reece2016ForecastingTO}
Andrew~G. Reece, Andrew~J. Reagan, Katharina L.~M. Lix, Peter~Sheridan Dodds,
  Christopher~M. Danforth, and Ellen~J. Langer. 2016.
\newblock Forecasting the onset and course of mental illness with twitter data.
\newblock \emph{Scientific Reports}, 7.

\bibitem[{Riglin et~al.(2020)Riglin, Leppert, Dardani, Thapar, Rice,
  O’Donovan, Smith, Stergiakouli, Tilling, and Thapar}]{Riglin2020ADHDAD}
Lucy Riglin, Beate Leppert, Christina Dardani, Ajay~K Thapar, Frances Rice,
  Michael~C. O’Donovan, George~Davey Smith, Evie Stergiakouli, Kate Tilling,
  and Anita Thapar. 2020.
\newblock Adhd and depression: investigating a causal explanation.
\newblock \emph{Psychological Medicine}, 51:1890 -- 1897.

\bibitem[{Sahito et~al.(2021)Sahito, Frank, and
  Pfahringer}]{Sahito2021BetterSF}
Attaullah Sahito, Eibe Frank, and Bernhard Pfahringer. 2021.
\newblock Better self-training for image classification through
  self-supervision.
\newblock \emph{ArXiv}, abs/2109.00778.

\bibitem[{Sampath et~al.(2023)Sampath, Durairaj, Chakravarthi, C,
  Shanmugavadivel, and Rahood}]{Sampath-depression-2023-overview}
Kayalvizhi Sampath, Thenmozhi Durairaj, Bharathi~Raja Chakravarthi,
  Jerin~Mahibha C, Kogilavani Shanmugavadivel, and Pratik~Anil Rahood. 2023.
\newblock Overview of the second shared task on detecting signs of depression
  from social media text.
\newblock In \emph{Proceedings of the Third Workshop on Language Technology for
  Equality, Diversity and Inclusion}, Varna, Bulgaria. Recent Advances in
  Natural Language Processing.

\bibitem[{Scudder(1965)}]{Scudder1965ProbabilityOE}
H.~J. Scudder. 1965.
\newblock Probability of error of some adaptive pattern-recognition machines.
\newblock \emph{IEEE Trans. Inf. Theory}, 11:363--371.

\bibitem[{T{\o}lb{\o}ll(2019)}]{Tlbll2019LinguisticFI}
Katrine~B{\o}nneland T{\o}lb{\o}ll. 2019.
\newblock Linguistic features in depression: a meta-analysis.

\bibitem[{Xie et~al.(2019)Xie, Hovy, Luong, and Le}]{Xie2019SelfTrainingWN}
Qizhe Xie, Eduard~H. Hovy, Minh-Thang Luong, and Quoc~V. Le. 2019.
\newblock Self-training with noisy student improves imagenet classification.
\newblock \emph{2020 IEEE/CVF Conference on Computer Vision and Pattern
  Recognition (CVPR)}, pages 10684--10695.

\bibitem[{Yazdavar et~al.(2017)Yazdavar, Al-Olimat, Ebrahimi, Bajaj, Banerjee,
  Thirunarayan, Pathak, and Sheth}]{Yazdavar2017SemiSupervisedAT}
Amir~Hossein Yazdavar, Hussein~S. Al-Olimat, Monireh Ebrahimi, Goonmeet Bajaj,
  Tanvi Banerjee, Krishnaprasad Thirunarayan, Jyotishman Pathak, and A.~Sheth.
  2017.
\newblock Semi-supervised approach to monitoring clinical depressive symptoms
  in social media.
\newblock \emph{Proceedings of the 2017 IEEE/ACM International Conference on
  Advances in Social Networks Analysis and Mining 2017}.

\bibitem[{Zhang et~al.(2022)Zhang, Schoene, Ji, and
  Ananiadou}]{Zhang2022NaturalLP}
Tianlin Zhang, Annika~Marie Schoene, Shaoxiong Ji, and Sophia Ananiadou. 2022.
\newblock Natural language processing applied to mental illness detection: a
  narrative review.
\newblock \emph{NPJ Digital Medicine}, 5.

\bibitem[{Zoph et~al.(2020)Zoph, Ghiasi, Lin, Cui, Liu, Cubuk, and
  Le}]{Zoph2020RethinkingPA}
Barret Zoph, Golnaz Ghiasi, Tsung-Yi Lin, Yin Cui, Hanxiao Liu, Ekin~Dogus
  Cubuk, and Quoc~V. Le. 2020.
\newblock Rethinking pre-training and self-training.
\newblock \emph{ArXiv}, abs/2006.06882.

\end{thebibliography}


\end{document}